\PassOptionsToPackage{table,xcdraw}{xcolor}
\documentclass[sigconf]{acmart}

\usepackage{xcolor}
\usepackage{soul}
\usepackage{booktabs}
\usepackage{multirow}
\usepackage[linesnumbered,ruled]{algorithm2e}
\usepackage{subcaption}
\usepackage{scalerel,graphicx,xparse}
\usepackage{rotating} 
\usepackage{lscape} 
\usepackage{enumitem}
\usepackage{threeparttable}

\AtBeginDocument{%
  \providecommand\BibTeX{{%
    \normalfont B\kern-0.5em{\scshape i\kern-0.25em b}\kern-0.8em\TeX}}}

\begin{document}

\title{Analyzing movies to predict their commercial viability for producers}


\author{Devendra Swami}
\affiliation{%
  \institution{University of Southern California}
  \city{Los Angeles}
  \state{California}
}
\email{dswami@usc.edu}

\author{Yash Phogat}
\affiliation{%
  \institution{University of Southern California}
  \city{Los Angeles}
  \state{California}
}
\email{phogat@usc.edu}

\author{Aadiraj Batlaw}
\affiliation{%
  \institution{University of California}
  \city{Berkeley}
  \state{California}
}
\email{batlaw33375@berkeley.edu}

\author{Ashwin Goyal}
\affiliation{%
  \institution{University of Southern California}
  \city{Los Angeles}
  \state{California}
}
\email{ashwingo@usc.edu}
\renewcommand{\shortauthors}{Swami et al.}

\begin{abstract}
Upon film premiere, a major form of speculation concerns the relative success of the film. This relativity is in particular regards to the film's original budget – as many a time have “big-budget blockbusters” been met with exceptional success as met with abject failure. So how does one predict the success of an upcoming film? In this paper, we explored a vast array of film data in an attempt to develop a model that could predict the expected return of an upcoming film. The approach to this development is as follows: First, we began with the MovieLens dataset \cite{movielens} having common movie attributes along with genome tags per each film. Genome tags give insight into what particular characteristics of the film are most salient. We then included additional features regarding film content, cast/crew, audience perception,  budget, and earnings from TMDB, IMDB, and Metacritic websites. Next, we performed exploratory data analysis and engineered a wide range of new features capturing historical information for the available features. Thereafter, we used singular value decomposition (SVD) for dimensionality reduction of the high dimensional features (ex. genome tags). Finally, we built a Random Forest Classifier and performed hyper-parameter tuning to optimize for model accuracy. A future application of our model could be seen in the film industry, allowing production companies to better predict the expected return of their projects based on their envisioned outline for their production procedure, thereby allowing them to revise their plan in an attempt to achieve optimal returns.
\end{abstract}

\maketitle

\section{Introduction}

The film industry is one that stands out in every aspect. It is a world of it's own. For this paper our motivation comes from the desire to provide a prediction model for producers to get an idea of the commercial viability of their proposed movie. Joe Swanberg said in 2016, "The only way you’re ever going to make any money is if you’re investing in your own movies". Before movie producers have to finalize their decision, they have to ensure that their investment is sound and understand how they see a return on that investment – this is where our model enters the movie world. Our main contributions in solving this problem can be summarized as follows:
\begin{itemize}
    \item We investigated a wide range of features that are likely to be associated with commercial success of a movie. Different from other available studies, we have incorporated many novel features such as publicity, release date and movie cast \& crew. We spent considerable time in feature engineering to better understand what factors make a movie financially lucrative.
    \item We extracted 11 different groups of features and built a random forest (RF) model to predict whether the return on investment (ROI) for the movie will be above or below median. After training RF, we have then identified the relative importance of each individual feature and groups of features. 
\end{itemize}

The remaining paper is structured as follows. In Section \ref{sec:researchmethod}, we describe our research methodology. In \ref{sec:evalresult}, we present our findings, followed by section \ref{sec:discussion}, where we discuss the relationship between few important features and ROI. In Section \ref{sec:threats}, we identify threats to validity and finally in section \ref{sec:conclusion}, we conclude the paper and suggest scope for future work.

\section{Research Methodology}
\label{sec:researchmethod}
\subsection{Problem Statement}
The current paper investigates the application of using a machine learning algorithm to provide better understanding of the features that could potentially affect the commercial success of a movie. The problem statement can be formally defined as follows

\textbf{\textit{Task: Movie Success Prediction}}: Given a movie $M$, predict whether $M$ will be a commercially successful movie.

\noindent While achieving this task we try to answer the research questions mentioned below.

\begin{itemize}[itemsep=0.5em]
    \item \textbf{RQ1}: How successful is the random forest algorithm in predicting whether a movie will be a commercial success in terms of ROI?
    \item \textbf{RQ2}: Which individual features and groups of features play the most important role in predicting ROI from movies?
\end{itemize}

\subsection{Features}
\label{sec:featureselection}

In this section, we discuss the features we have considered in this study. The features were selected based on what features are generally used for performing analysis of movies industry, and other novel features that are publicly available and can be extracted easily using accessible tools.   

Table \ref{tab:summaryoffeature} shows the summary of features considered in this study. We categorized our selected features into $11$ groups based on the characteristics of a feature within the group. Each feature group is summarized below. 

\begin{table*}[ht]
    \centering
    \caption{Movie features likely to affect the commercial success of a movie}
    \label{tab:summaryoffeature}
    \setlength{\tabcolsep}{12pt}
    \renewcommand{\arraystretch}{1.3}
    \resizebox{0.98\textwidth}{!}{%
    \begin{threeparttable}
\begin{tabular}{llll} 
\toprule
\textbf{\#} & \textbf{Group}          & \textbf{Feature}           & \textbf{Description}                                                                               \\ 
\hline
\multirow{6}{*}{1} &\multirow{6}{*}{Content}      & is\_adult                & Whether the movie is only suitable for adults or not.             \\ 
\cline{3-4}
          &                & is\_english               & Whether the primary language used in the movie is English or not.                    \\ 
\cline{3-4}
          &                 & languages\_count              & The total number of languages used in the movie.                                                   \\ 
\cline{3-4}
          &                 & movie\_runtime         & The total length of the movie, in minutes.                 \\ 
\cline{3-4}
          &                 & genome\_tags \cite{genome_paper}         & Features extracted from movie content.                 \\
\cline{3-4}
          &                 & movie\_genres         & The different genres under which the movie can be placed into.                 \\

\hline

\multirow{4}{*}{2} & \multirow{4}{*}{Publicity} & is\_collection    & Whether the movie is part of a collection or series.       \\ 
\cline{3-4}
            &              & is\_homepage           & Whether the movie has a homepage or not.        \\ 
\cline{3-4}
            &              & is\_tagline           & Whether the movie has an associated tagline or not.              \\ 
\cline{3-4}
            &              & keywords\_count         & Number of frequently used keywords that can be assigned to the movie.            \\ 
\hline

\multirow{6}{*}{3} & \multirow{6}{*}{Audience Perception$\dagger$} & popularity               & The movie popularity score provided by TMDB.                                        \\ 

\cline{3-4}
              &               & vote\_average                 & The average approval score for the movie, based on total votes at TMDB.                                        \\ 
\cline{3-4}
              &               & vote\_count                & The total number of votes provided for the movie at TMDB.                                       \\ 
\cline{3-4}
              &               & metacritic\_score                 & The average approval movie score at metacritic.                                        \\ 
\cline{3-4}
              &               & imdb\_rating                 & The average approval movie score at IMDB.                                        \\ 
\cline{3-4}
              &               & imdb\_votes                 & The total number of votes provided for the movie at IMDB.                                        \\ 
              
\hline
\multirow{4}{*}{4} & \multirow{4}{*}{Release Date} & release\_month               & The month in which the movie is released.                                        \\ 

\cline{3-4}
              &               & movies\_per\_month                 & The total number of movies released in the same month as the given movie.                                        \\ 
\cline{3-4}
              &               & budget\_fraction                & The budget of the movie in proportion to the cumulative budget of all movies of that month.                                        \\ 
\cline{3-4}
              &               & movie\_expense\_score                 & The budget of the movie in proportion to the average budget of all movies of that month.                                        \\ 
\hline

5 & Finance                     & time\_discounted\_budget                & The time discounted value for budget used to produce movie.                                                                 \\ 
\hline

6 & Production House                     & production\_house\_embedding                & Calculated from performance of recent movies produced by the same production houses.                                                                 \\ 
\hline

7 & Writers                     & writers\_embedding                & Calculated from performance of recent movies written by the same writers.                                                                 \\ 
\hline

8 & Directors                     & directors\_embedding                & Calculated from performance of recent movies directed by the same directors.                                                                 \\ 
\hline

9 & Producers                     & producers\_embedding                & Calculated from performance of recent movies produced by the same producers.                                                                 \\ 
\hline

10 & Main Cast                     & main\_cast\_embedding                & Calculated from performance of recent movies where members from the main cast featured.                                                                \\ 
\hline 

\multirow{3}{*}{11} & \multirow{3}{*}{Support Staff} & female\_count               & The total number of females in the movie cast.                                        \\ 

\cline{3-4}
              &               & male\_count                 & The total number of males in the movie cast.                                        \\ 
\cline{3-4}
              &               & crew\_length                 & The total number of people in the movie crew.                                        \\ 

\bottomrule
\end{tabular}
\begin{tablenotes}
    \item $\dagger$: Not used in prediction model as that information is not available before movie release.
\end{tablenotes}
\end{threeparttable}
    }
\end{table*}

\subsubsection{\textbf{Content}}
In this group, we consider \textit{is\_adult}, \textit{is\_english}, \textit{languages\_count}, \textit{runtime}, \textit{genome} and \textit{genre}. The content of the film is important as film watchers typically have a distinct preference for these categories (ex. English speakers typically prefer movies in English), and therefore film content allows us to narrow down on the target audience.

\subsubsection{\textbf{Publicity}}Several features were extracted to measure publicity efforts, like \textit{is\_collection}, \textit{is\_homepage}, \textit{is\_tagline} and \textit{keywords\_count}. Publicity is an important factor, as the marketing and accessibility of the film has a direct impact on the number of people that hear about the movie, and, by extension, the number of people that watch the movie (today in particular, films part of series tend to out perform other movies on the box office). 

\subsubsection{\textbf{Audience Perception}} In this group, we examined the features including \textit{popularity}, \textit{vote\_average}, \textit{vote\_count}, \textit{metacritic\_score}, \textit{imdb\_rating}, \textit{imdb\_votes}. Naturally, audience perception has a huge effect on the success of movies. Films with high ratings tend to outperform movies with lower ratings. However, since this information is usually unavailable until after the film is released, we did not include this in our prediction model.

\subsubsection{\textbf{Release Date}}
In this group, we examined features like \textit{release\_month}, \textit{movies\_per\_month}, \textit{budget\_fraction}, \textit{expense\_score}. Several studies have indicated that the competition a film meets at the time of release has a major impact on the financial success of the film. This feature group strives to examine this effect.

\subsubsection{\textbf{Finance}} 
This group comprises solely of the feature,

\textit{time\_discounted\_budget}, which is the discounted value of the budget used to produce the movie. Typically films with larger budgets tend to outperform lower budgeted movies in terms of revenue.

\subsubsection{\textbf{Production House}}
This group only includes the feature \textit{production\_house\_embedding}. This is calculated from average performance of recent movies produced by the same production houses. Better, more established production houses are usually able to hire more renowned directors, writers, and stars, which, in addition to working with larger budgets, are factors that typically lead to box office success.

\subsubsection{\textbf{Writers}}
This group comprises solely of the feature  \textit{writers\_embedding}, which was calculated from average performance of recent movies written by the same writers. Typically, writers who have written for successful films tend to write more successful films.

\subsubsection{\textbf{Directors}}
This group comprises solely of the feature  \textit{directors\_embedding}, which was calculated from average performance of recent movies made by the same directors. Typically, directors who have directed successful films tend to direct more successful films.

\subsubsection{\textbf{Producers}}
This group comprises solely of the feature  \textit{producers\_embedding}, which was calculated from average performance of recent movies made by the same producers. Typically, producers who have produced successful films tend to produce more successful films.

\subsubsection{\textbf{Main Cast}}
This group comprises solely of the feature  \textit{main\_cast\_embedding}, which is calculated from average performance of recent movies where members from the main cast featured. Several studies have shown that the star(s) of the film have a major effect on its success.

\subsubsection{\textbf{Support Staff}}
This group comprises of the features \textit{female\_count}, \textit{male\_count}, and \textit{crew\_length}. Some studies indicate the gender and size of the cast of a film can have an effect on its success.

\subsection{Data Collection}

We obtained most of the data from the Kaggle dataset "The Movies Dataset",\cite{kaggleTMDB} which provided us with data on the cast, crew, and general metadata on a large subset of films. In a dataset provided by the Data Open competition community, we obtained the genome tags, which we merged with our metadata dataset. We obtained further features by making API calls to the TMDB (The Movie Database), as well as scraping Metacritic scores from "IMDB: All U.S. Released Movies: 1972-2016". Initially, our merged dataset included slightly over 13K rows as we only had genome information for these many movies. We then decided to focus on movies after the color revolution (i.e. after 1965), and got rid of rows that had their budget values or revenue values set to 0. (There was forethought to interpolate these errors, however, as budget and revenue were such an important factor in determining rate of return, we concluded any method of interpolation would largely bias our model). This brought our total number of rows to 5,426.

\subsection{Machine Learning Algorithm}

We first thought of using regression as our machine learning algorithm due to the fact that the ROI values are continuous. However, our predictions from regression were rather inaccurate in many cases. Moreover, rather than being interested in knowing exact values, movie producers might be more inclined to know whether their movie will potentially perform well or otherwise. Thus, we have decided to predict above or below median ROI \textit{(median calculated on training data)} rather than precise ROI values. This way, we have converted our original hard-to-learn regression problem into an easy binary classification task. 

We have deployed random forest (RF) algorithm to perform our classification task since it is one of the most successful non-linear machine learning algorithm. It considers numerous decision trees trained over random sample of the training data. In addition, for splitting nodes it randomly selects a subset of features. Finally, the decision is carried out by taking average of predictions from each decision tree. As a result, the random forest algorithm is less prone to over-fitting. Moreover, it can take numerical or categorical variables as input and even does not require feature scaling. For these reasons, we chose the random forest algorithm for performing the classification task. The dataset is split into approx 80\% training and 20\% test data with movies earlier than 2011 are part of train data and from 2011 and on-wards comprising test set. 

\textbf{Dimensionality reduction:} Due to presence of high dimensional sparse features like genome tags and genre information (after one hot encoding) in the dataset, we have used singular value decomposition (SVD) to reduce their number of dimensions while keeping most of the variance in data intact. In addition to that, we have also removed highly correlated features from our dataset. We have identified highly correlated pair of features, i.e. pairs having absolute correlation value of 0.75 or higher from Spearman's correlation. Among these identified pairs, features having a lower mutual information \cite{kraskov2004estimating} are dropped. It is done to remove features that do not provide any additional information and can help in reducing size of the dataset to speed up our training process.

\textbf{Hyperparameters optimization:} For obtaining optimal hyperparameters for our RF model, we have started by preparing a grid search space. For \textbf{n\_estimators}, we considered values from 100 to 1000 with step size of 100. Similarly, integral multiple of $\sqrt{\#features}$ to \#features for \textbf{max\_features}, values from 10 to 100 with step size of 10 for \textbf{max\_depth} is used. Values of 0.01, 0.03, and 0.05 is considered for \textbf{min\_samples\_split}, and 1, 3, and 5 for \textbf{min\_samples\_leaf}. We have always set \textbf{bootstrap} to \textit{True}.

Due to grid search space of $12600$ ($10\cdot14\cdot10\cdot3\cdot3$) being computationally expensive, we have performed randomized search over uniformly drawn samples from it. We have used $100$ iterations through $4-$fold cross validation to identify the optimal hyperparameters. We report the hyperparameters obtained from this procedure in Table 
\ref{table:model_params}.
\begin{table}[htbp]
\caption{Parameters of RF model tuned over validation set}
\begin{tabular}{|l|c|}
\toprule
\textbf{Parameters} & \textbf{Optimal Value} \\ \midrule
n\_estimators          & 500             \\
max\_features          & 14             \\
max\_depth         & 40             \\
min\_samples\_split          & 0.05             \\
min\_samples\_leaf         & 5             \\ \bottomrule
\end{tabular}
\label{table:model_params}
\end{table}

\subsection{Model Evaluation}

By virtue of the way we formulate our classification task, our dataset is perfectly balanced with two classes of equal importance. Therefore, \textit{Accuracy} is initially thought to be a suitable evaluation metric. However, accuracy is sensitive to selection of threshold for converting predicted probability to the most likely output class. Different threshold can yield different accuracy score for the same ML predictions and thus makes it difficult to do performance comparison across multiple ML algorithms. 

In order to overcome this hurdle, we have selected the Area Under the Receiver Operating Characteristic (ROC) Curve to evaluate the performance on our task. We use the acronym \textit{AUC} to denote this metric. The ROC curve is a plot between true positive rate and false positive rate obtained by deploying different thresholds to assign positives. The area under this curve (AUC) is then used to compare the performance of a model over our binary classification task. 

Further, we have compared the performance of the random forest algorithm with the random baseline approach. In \textbf{Random baseline}, we randomly assign below or above median ROI to each movie in the test set. An absence of significant improvement in performance of RF model over baseline will imply that the selected features have no association with commercial success of the movie. A perfect classification model will yield an AUC value of $1.00$, whereas a random model will have value around $0.50$. In general, a higher AUC value indicates a better model.

\subsection{Feature Importance Analysis}
\label{sec:importance}

We have used permutation feature importance technique proposed by Breiman \cite{breiman2001random} to measure the importance of an individual feature and a group of features. The key idea behind this method is that random permutation of important features should lead to a substantial deterioration of model performance. This decrease in performance metric is referred as the \textbf{Importance Value (IV)} and a higher value indicates that the feature is more important. 

Similarly, the group importance is calculated by permuting all features in a group together. It is important to not here that we can safely employ the permutation feature importance technique \cite{strobl2008conditional} since none of the features are highly correlated in our dataset as we have already removed highly correlated features.

\section{Results}
\label{sec:evalresult}

\subsection{RQ1: How successful is the random forest algorithm in predicting whether a movie will be a commercial success in terms of ROI?}

\begin{figure}[ht]
    \centering
    \includegraphics[width=0.9\columnwidth]{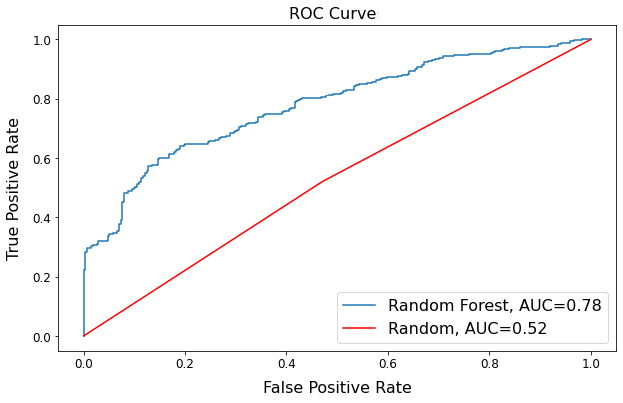}
    \caption{The ROC curve of the random forest algorithm that uses the selected features, and the baseline for our dataset.}
    \label{fig:roc_curves_all}
\end{figure}

Figure \ref{fig:roc_curves_all} shows the ROC curves of the baselines and the random forest algorithm . The AUC for the random forest algorithms is 0.78. Regarding the baseline, \textit{Random} achieved expected AUC scores of around 0.500. A higher AUC score of RF model suggests that a solution that uses feature engineering and machine learning would have the potential to help movie producers and analysts to better estimate the expected ROI for upcoming movies.

\subsection{RQ2: Which individual features and groups of features play the most important role in predicting ROI from movies?}

\begin{table}[ht]
\caption{Importance of top 15 individual features}
\begin{tabular}{lc}
\hline
\textbf{Feature Name} & \textbf{Information Value (IV)} \\ \hline
is\_collection          & 0.010             \\
genome\_0          & 0.005             \\
keywords\_count         & 0.004             \\
genome\_2          & 0.004             \\
genome\_13         & 0.004             \\
movies\_per\_month          & 0.003             \\
male\_count         & 0.002             \\
genome\_3          & 0.002             \\
genome\_5          & 0.002             \\
genome\_12          & 0.002             \\
is\_homepage          & 0.001             \\
time\_discounted\_budget          & 0.001             \\
female\_count          & 0.001             \\
genome\_1         & 0.001             \\
genome\_10          & 0.001         \\ \hline   
\end{tabular}
\label{tab:featureimp}
\end{table}

Table \ref{tab:featureimp} shows the top 15 most important features in predicting movies' ROI. We also observe that \textit{is\_collection} is among the most important feature. This makes intuitive sense, as sequels, spin-offs, and films extensions of a thematic universe tend to outperform movies without such characteristics. We also notice that the genome features (which as one might recall describes a film by its most salient characteristics) tend to offer valuable information. \textit{key\_word\_count} offers similar information to the genome feature, and it also having a high information valuable implies that the characteristics of films are a key insight for determining the success of a film. The last noticeable observation in this table would be \textit{movies\_per\_month}, which implies that competition upon a film's release date is indeed a major contributor in determining the success of a film.

\begin{table}[htbp]
\caption{Importance of top 5 feature groups}
\begin{tabular}{{|l|c|}}
\toprule
\textbf{Feature Group} & \textbf{Information Value (IV)} \\ \midrule
Content          & 0.047             \\
Main Cast          & 0.031             \\
Publicity          & 0.021             \\
Writers          & 0.017             \\
Production House         & 0.016             \\ \bottomrule
\end{tabular}
\label{tab:feature_group_imp}
\end{table}

Table \ref{tab:feature_group_imp} shows the top 5 most important feature groups. Here we observe that \textit{Content} has the highest information value. This makes sense, as \textit{genome\_tags} is included as a feature in this category. With \textit{Main Cast} coming in second in IV, we reaffirm our hypothesis that the stars of a film have a major impact on its success.

\section{Discussion}
\label{sec:discussion}
Rather than solely rely on feature importance values discussed in previous section, we also want to investigate the causality and determine the direction of influence of important features on ROI values. Thus, in this section, we specifically explore the relationship between the key features and return. However, we note that this is a uni-variate analysis which has the shortcoming of not considering multivariate interactions.

\begin{figure}[ht]
    \centering
    \includegraphics[width=1.0\columnwidth]{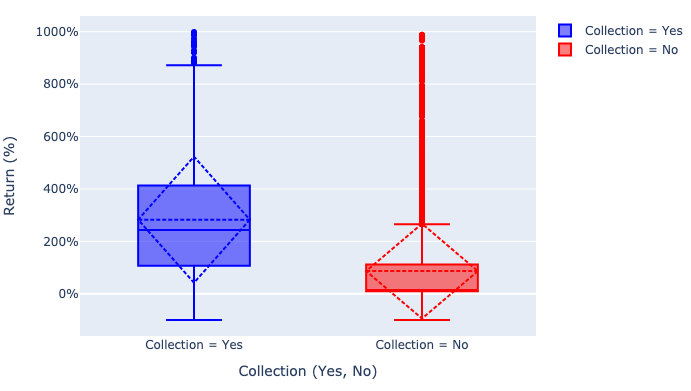}
    \caption{Plot between Is\_collection and Return}
    \label{fig:collection_vs_return}
\end{figure}

Figure \ref{fig:collection_vs_return} shows that there appears to be a reasonable difference in the return on investment of films that are a part of a collection and those that are not. Movies that are part of a collection/series tend to offer higher ROI than those not part of any collection.

\begin{figure}[t]
    \centering
    \includegraphics[width=1.0\columnwidth]{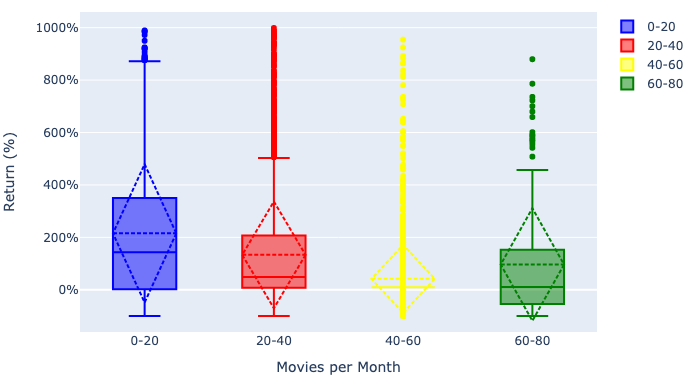}
    \caption{Plot between Movies per month and Return}
    \label{fig:movies_per_month_vs_return}
\end{figure}

Figure \ref{fig:movies_per_month_vs_return} shows that there appears to be a fairly reasonable difference between the return on investment for movies released with different ranges of number of other movies released in that month. It appears that for the most part, the fewer movies released in the month of the released date of the film of interest, the higher the rate of return. It is important to note that there are large outliers for all these categories, signalling that some films across categories performed exceptionally well regardless of the number of films released in the same month of the film's release date. 

\begin{figure}[bp]
    \centering
    \includegraphics[width=1.0\columnwidth]{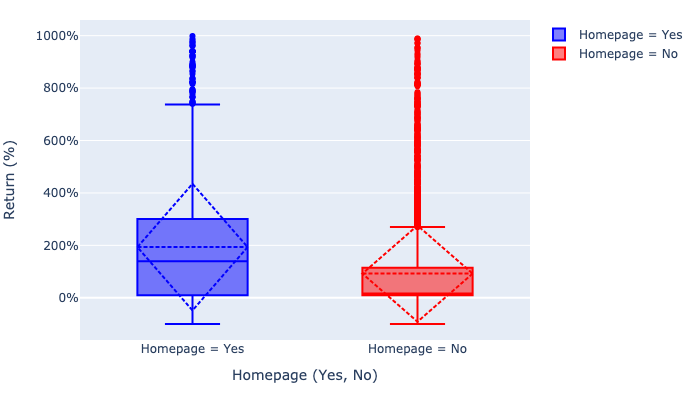}
    \caption{Plot between Is\_homepage and Return}
    \label{fig:homepage_vs_return}
\end{figure}

Figure \ref{fig:homepage_vs_return} explores if the existence of a homepage for a particular film is associated with a high return on investment. From this figure, it appears there is an association, but it does not seem to be particularly significant for the univariate case as we see there confidence intervals over lap.

\begin{figure}[htbp]
    \centering
    \includegraphics[width=1.0\columnwidth]{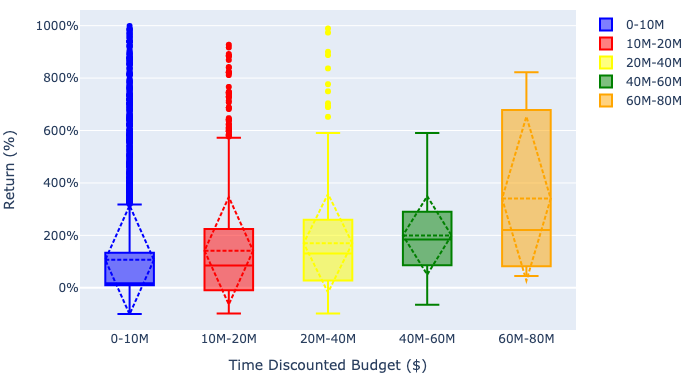}
    \caption{Plot between Time Discounted Budget and Return}
    \label{fig:budget_vs_return}
\end{figure}

Figure \ref{fig:budget_vs_return} compares the return on investment for different categories of budget. From here we see that there films produced with larger budgets tend to have a higher return on investment. 

\section{Threats to Validity}
\label{sec:threats}
We have listed out various threats to the validity and limitations of our study in this section.

\textbf{Feature Selection Bias:}  The results of current study can be affected by the features we have considered, limited by our imagination. Thus, others employing a different set of features may obtain different experimental results. There are few features that we have considered for our prediction but could not incorporate yet for example if the movie was nominated for an academy award, how well connected was the production team with the world? Yes, we have included popularity features but the broader question remains, how was the budget managed? How much was allocated to marketing and campaigning?. Thus, including more features in consultation with industry experts is part of our future plan.

\smallskip
\textbf{Tool and Method Reliability:} Although we have used standard tools and methods suitable for such studies, it is still possible that inaccuracies exist in them that we did not take into account. For example, we assumed that genome tags are perfectly computed and provide a better representation of the movie content. However, being derived from ML algorithm, these tags can prove to be inaccurate in some cases.  

\smallskip

\textbf{External validity:} We agree that there are several limitations that could potentially influence the generalizability of our findings. First, the movie sampling problem; to elaborate, we realised that times have changed since early 1920's and hence we would need to consider only a limited time-frame before the targeted period, this led to reduction in data points and hence reduction in validity checks that we could perform. 

Further, only one machine learning algorithm is used, namely Random Forest. We considered using neural networks but due a paucity of data points our model would have been flawed.

\section{Conclusions and Future Work}
\label{sec:conclusion}

In this paper, we sought to answer the question how might production companies predict future return of upcoming cinematic projects, attempting to help filmmakers better understand the commercial viability of their undertakings. To achieve this, we extracted features in the categories of Content, Publicity, Release Date, Finance, Production House, Writers, Directors, Producers, Main Cast, and Support Staff. Finally, we built a random forest classification model, which, after thorough hyperparameter tuning, was able to successfully predict the return of the films with an AUC score of \text{78\%}. 

\smallskip
\textbf{Future Work:} The top priority is adding more features and expanding our data-set to include more data-points for available features. This includes web-scrapping social media websites to find hidden connections and deep networks that were previously unknown, for example, constructing a social media connectivity graph to better identify popularity and inner circles of the production team which in-turn provides better predictions for our core problems. (Social circles identify how powerful a single member's presence is - better the presence, more the return). Once we have done that, we would like to concentrate our efforts in employing other learning algorithms to obtain new models that would compare or ensemble with our current model. One such algorithm can be neural networks as we may discover additional real world relationships that are currently unknown. This, of course is possible once we have expanded our data-set sufficiently.

\bibliographystyle{acm}
\bibliography{main.bib}

\end{document}